\documentclass[final]{cvpr}

\usepackage{times}
\usepackage{epsfig}
\usepackage{graphicx}
\usepackage{amsmath}
\usepackage{amssymb}
\usepackage{mathtools}
\usepackage{booktabs} 
\usepackage{enumitem}
\usepackage{comment}
\usepackage{bm}
\usepackage[accsupp]{axessibility}  
\newcommand{\myparagraph}[1]{\textbf{#1}}

\usepackage[pagebackref=true,breaklinks=true,colorlinks,bookmarks=false]{hyperref}

\def\cvprPaperID{9293} 
\def\confYear{CVPR 2022}

\newcommand\blfootnote[1]{%
  \begingroup
  \renewcommand\thefootnote{}\footnote{#1}%
  \addtocounter{footnote}{-1}%
  \endgroup
}

\usepackage{cleveref}

\begin{document}

\title{\Large \bf RSTT: Real-time Spatial Temporal Transformer for Space-Time Video Super-Resolution}




\author{Zhicheng Geng$^{1*}$
\and
Luming Liang$^{2*\dagger}$
\and
Tianyu Ding$^2$
\and
Ilya Zharkov$^2$\\
}

\date{$^1$University of Texas, Austin \qquad  $^2$Microsoft\\
{\tt\small zhichenggeng@utexas.edu},
{\tt\small \{lulian,tianyuding,zharkov\}@microsoft.com}
}

\maketitle

\begin{abstract}
   Space-time video super-resolution (STVSR) is the task of interpolating videos with both Low Frame Rate (LFR) and Low Resolution (LR) to produce High-Frame-Rate (HFR) and also High-Resolution (HR) counterparts. The existing methods based on Convolutional Neural Network~(CNN) succeed in achieving visually satisfied results while suffer from slow inference speed due to their heavy architectures. We propose to resolve this issue by using a spatial-temporal transformer that naturally incorporates the spatial and temporal super resolution modules into a single model. Unlike CNN-based methods, we do not explicitly use separated building blocks for temporal interpolations and spatial super-resolutions; instead, we only use a single end-to-end transformer architecture. Specifically,  a reusable dictionary is built by encoders based on the input LFR and LR frames, which is then utilized in the decoder part to synthesize the HFR and HR frames. Compared with the state-of-the-art TMNet~\cite{xu2021temporal}, our network is $60\%$ smaller (4.5M vs 12.3M parameters) and $80\%$ faster (26.2fps vs 14.3fps on $720\times576$ frames) without sacrificing much performance. {The source code is available at \url{https://github.com/llmpass/RSTT}}.
   
\end{abstract}

\section{Introduction}

\blfootnote{$^*$Equal contribution. This work was done when Zhicheng Geng was an intern at Applied Sciences Group, Microsoft. $^\dagger$ Corresponding author, llmpass@gmail.com.
}

\begin{figure}[t]
\begin{center}
   \ \ \includegraphics[width=0.95\linewidth]{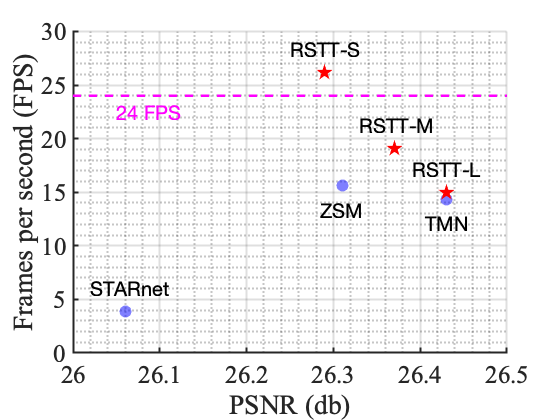}\\
   \ \ \includegraphics[width=0.95\linewidth]{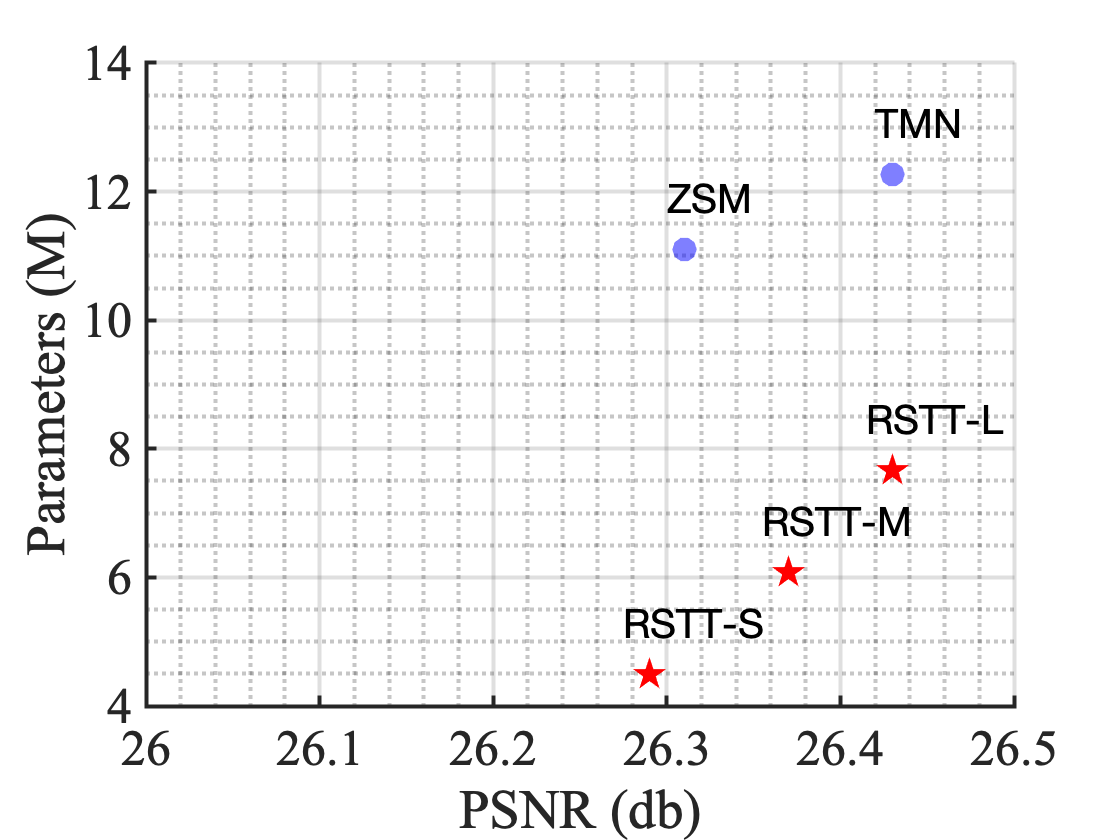}
\end{center}
   \caption{\textbf{Performance of RSTT on Vid4 dataset~\cite{liu2011bayesian} using small (S), medium (M) and large (L) architectures compared to other baseline models.} In the top, we plot FPS versus PSNR. Note that 24 FPS is the standard cinematic frame rate~\cite{tag2016eye}. In the bottom, we plot the number of parameters (in millions) versus PSNR. We omit the STARnet here since it is significantly larger than others while performs the worst; see~\Cref{QuantEva} for details. }
\label{fig:long}
\label{fig:onecol}
\vspace{-.2in}
\end{figure}

Space-time video super-resolution~(STVSR) refers to the task of simultaneously increasing spatial and temporal resolutions of low-frame-rate~(LFR) and low-resolution~(LR) videos, which appears in a wide variety of multimedia applications such as video compression~\cite{rippel2019learned}, video streaming~\cite{wu2001streaming}, video conferencing~\cite{turletti1996videoconferencing} and so on. In the stage of deployment, many of them have stringent requirements for the computational efficiency, and only LFR and LR frames can be transferred in real-time. STVSR becomes a natural remedy in this scenario for recovering the high-frame-rate (HFR) and high-resolution (HR) videos.
However, the performance of existing STVSR approaches are far from being real-time, and a fast method without sacrificing much visual quality is crucial for practical applications. 


The success of traditional STVSR approaches usually relies on strong illumination consistency assumptions~\cite{mudenagudi2010space,shechtman2005space}, which can be easily violated in real world dynamic patterns. Ever since the era of deep neural network (DNN), convolutional neural network (CNN) exhibits promising results in many video restoration tasks, \eg, video denoising~\cite{tassano2020fastdvdnet,claus2019videnn}, video inpainting~\cite{kim2019deep,wang2019video}, video super-resolution~(VSR)~\cite{liu2011bayesian,JoCVPR2018,tian2018tdan,wang2019edvr,RBPN2019} and video frame interpolation~(VFI)~\cite{jiang2018super,bao2019depth,cheng2020multiple,niklaus2017video_sepcov,niklaus2017video,niklaus2018context,niklaus2020softmax,ding2021cdfi}. One straightforward way to tackle the STVSR problem is that by treating it as a composite task of VFI and VSR one can sequentially apply VFI, \eg, SepConv \cite{niklaus2017video_sepcov}, DAIN \cite{bao2019depth}, CDFI \cite{ding2021cdfi}, and VSR, \eg, DUF \cite{JoCVPR2018}, RBPN \cite{RBPN2019}, EDVR \cite{wang2019edvr}, on the input LFR and LR video. Nevertheless, this simple strategy has two major drawbacks:
first, it fails to utilize the inner connection between the temporal interpolation and spatial super-resolution~\cite{xiang2020zooming,xu2021temporal}; second, both VFI and VSR models need to extract and utilize features from nearby LR frames, which results in duplication of work. Additionally, such two-stage models usually suffer from slow inference speed due to the large amount of parameters, hence prohibits from being deployed on real-time applications. 

To alleviate the above problems, recent learning based methods train a single end-to-end model~\cite{STAR2020,xiang2020zooming,xiang2021zooming,xu2021temporal}, where features  are extracted from the input LFR and LR frames only once and then are upsampled in time and space sequentially inside the network. However, researches in this line still consist of two sub-modules: a temporal interpolation network, \eg, Deformable ConvLSTM~\cite{xiang2020zooming,xiang2021zooming} and Temporal Modulation Block~\cite{xu2021temporal}, and a spatial reconstruction network, \eg, residual blocks used in both Zooming SloMo~\cite{xiang2020zooming,xiang2021zooming} and TMnet~\cite{xu2021temporal}. One natural question to ask is that whether we can have a \emph{holistic design} such that it increases the spatial and temporal resolutions simultaneously without separating out the two tasks.

In this paper, we propose a single \emph{spatial temporal transformer} that incorporates the temporal interpolation and spatial super resolution modules for the STVSR task. This approach leads to a much smaller network compared with the existing methods, and is able to achieve a real-time inference speed without sacrificing much performance. Specifically, we make the following contributions:

\begin{itemize}[leftmargin=*]
    \item We propose a Real-time Spatial Temporal Transformer~(RSTT) to increase the spatial and temporal resolutions without explicitly modeling it as two sub-tasks. To the best of our knowledge, this is the first time that a transformer is utilized to solve the STVSR problem.
    
    \item Inside RSTT, we design a cascaded UNet-style architecture to effectively incorporate all the spatial and temporal information for synthesizing HFR and HR videos. In particular, the encoder part of RSTT builds dictionaries at multi-resolutions, which are then queried in the decoder part for directly reconstructing HFR and HR frames. 
    
    \item We propose three RSTT models with different number of encoder and decoder pairs, resulting in small (S), medium (M) and larger (L) architectures. Experiments show that RSTT is significantly smaller and faster than the state-of-the-art STVSR methods while maintains similar performance: 
    {(i)} RSTT-L performs similarly as TMNet~\cite{xu2021temporal} with $40\%$ less parameters, RSTT-M outperforms Zooming SlowMo~\cite{xiang2020zooming} with $50\%$ less parameters and RSTT-S outperforms STARNet \cite{STAR2020} with $96\%$ less parameters;
    {(ii)} RSTT-S achieves a frame rate of more than 24 per second (the standard cinematic frame rate) on $720\times 576$ frames. It achieves the performance of Zooming SlowMo~\cite{xiang2020zooming} with a $75\%$ speedup, and outperforms STARNet~\cite{STAR2020} with around $700\%$ speedup (see~\Cref{fig:onecol}). 
\end{itemize}

\section{Related work}
\subsection{Video frame interpolation (VFI)}

VFI aims at synthesizing an intermediate frame given two consecutive frames in a video sequence, hence the temporal resolution is increased. Conventionally, it is formulated as an image sequence estimation problem, \eg, the path-based~\cite{mahajan2009moving} and phase-based approach~\cite{meyer2018phasenet,meyer2015phase}, while it fails in scenes with fast movements or complex image details. CNN-based VFI methods can be roughly categorized into three types: flow-based, kernel-based and deformable-convolution-based. Flow-based methods~\cite{liu2017video,jiang2018super,park2020bmbc,xue2019video,yuan2019zoom,niklaus2018context,niklaus2020softmax} perform VFI by estimating optical flow between frames and then warping input frames with the estimated flow to synthesize the missing ones. Instead of using only pixel-wise information for interpolation, kernel-based methods~\cite{niklaus2017video,niklaus2017video_sepcov,bao2019depth,bao2019memc} propose to synthesize the image by convolving over local patches around each output pixel, which largely preserves the local textual details. Recently,  deformable-convolution-based methods~\cite{lee2020adacof,shi2020video,cheng2020video,cheng2020multiple,ding2021cdfi} combine flow-based and kernel-based methods
by taking the advantages of flexible spatial sampling  introduced by deformable convolution~\cite{dai2017deformable,zhu2019deformable}. This key improvement accommodates to both larger motions and complex textures.

\subsection{Video super resolution (VSR)}
VSR aims to recover HR video sequences from LR ones, hence the spatial resolution is increased. Most deep learning based VSR methods~\cite{liu2020video} adopt the strategy of fusing spatial features from multiple aligned frames (or features), which highly depends on the quality of the alignment. Earlier methods~\cite{caballero2017real,tao2017detail,sajjadi2018frame, wang2018learning,xue2019video} utilize optical flow to perform explicit temporal frames alignment. However, the computation of optical flow can be expensive and the estimated flow can be inaccurate. In parallel, TDAN~\cite{tian2018tdan} introduces deformable convolution to implicitly align temporal features and achieves impressive performance, while EDVR~\cite{wang2019edvr} incorporates deformable convolution into a multi-scale module to further improve the feature alignments.

\subsection{Space-time video super-resolution (STVSR)}

The goal of STVSR is to increase both spatial and temporal resolutions of LFR and LR videos. Dating back two decades, \cite{shechtman2002increasing} performs super-resolution simultaneously in time and space by modeling the dynamic scene as 3D representation. However, it requires input sequences of several different space-time resolutions to construct a new one. 
Due to the recent success of CNN, \cite{STAR2020} proposes an end-to-end network STARnet to increase the spatial resolution and frame rate by jointly learning spatial and temporal contexts. Xiang et. al~\cite{xiang2020zooming} propose a one-stage framework, named Zooming SlowMo, by firstly interpolating temporal features using deformable convolution and then fusing mutli-frame features through deformable ConvLSTM. Later, Xu et. al~\cite{xu2021temporal} incorporate temporal modulation block into Zooming SlowMo~\cite{xiang2020zooming}, named TMNet, so that the model is able to interpolate arbitrary intermediate frames and achieves better visual consistencies in between resulting frames. Nevertheless, these approaches either explicitly or implicitly treat the STVSR problem as a combination of VFI and VSR by designing separate modules for the sub-tasks, which is computationally expensive. Zhou et al. point out in a more recent work \cite{Zhou2021ACMMM} that VFI and VSR mutually benefit each other. They present a network that cyclically uses the intermediate results generated by one task to improve another and vice versa. While achieving better performance, this idea results in a relatively larger network (about three times larger than Zooming SlowMo~\cite{xiang2020zooming}). Our work also makes use of such mutual benefits from VFI and VSR, while avoids the repeated feature computations, and thus ends in a light-weight design.

\subsection{Vision transformer}

Transformer~\cite{vaswani2017attention} is a dominant architecture in Natural Language Processing (NLP) and achieves state-of-the-art performance in 
various tasks~\cite{devlin2018bert,brown2020language,xu2021bert}.
Recently, transformers gain popularity in computer vision field.
The pioneering Vision Transformer (ViT)~\cite{dosovitskiy2021an} computes attentions between  flattened image patches to solve image classification problems and outperforms CNN-based methods to a large extent. Due to transformer's strong ability of learning long dependencies between different image regions, follow-up work using variants of ViT refreshes the state-of-the-art results in many applications, such as segmentation~\cite{strudel2021segmenter,zheng2021rethinking}, object detection~\cite{yang2021focal,xu2021end} and depth estimation~\cite{ranftl2021vision}. In the meantime, Liu et al. \cite{liu2021swin} propose a novel transformer-based backbone for vision tasks, \ie, Shifted window (Swin) Transformer, to reduce computational complexity by restricting the attention computations inside local and later shifted local windows.
Thereafter, \cite{wang2021uformer} proposes a U-shape network based on Swin Transformer for general image restoration. SwinIR~\cite{liang2021swinir} tackles the image restoration task using Swin Transformer and introduces residual Swin Transformer blocks. 
In this work, we also use Swin Transformer as basic building blocks to extract local information. However, instead of building dictionaries and queries from identical single frames~\cite{liu2021swin,chen2021pre,wang2021uformer,liang2021swinir}, we compute window and shifted window attentions from multiple input frames, which are then utilized to build reusable dictionaries to synthesize multiple output frames at once. We emphasize that this design is the key that leads to the acceleration of inference time and reduction of model size. 


\begin{figure*}[h]
\begin{center}
\includegraphics[width=0.99\textwidth]{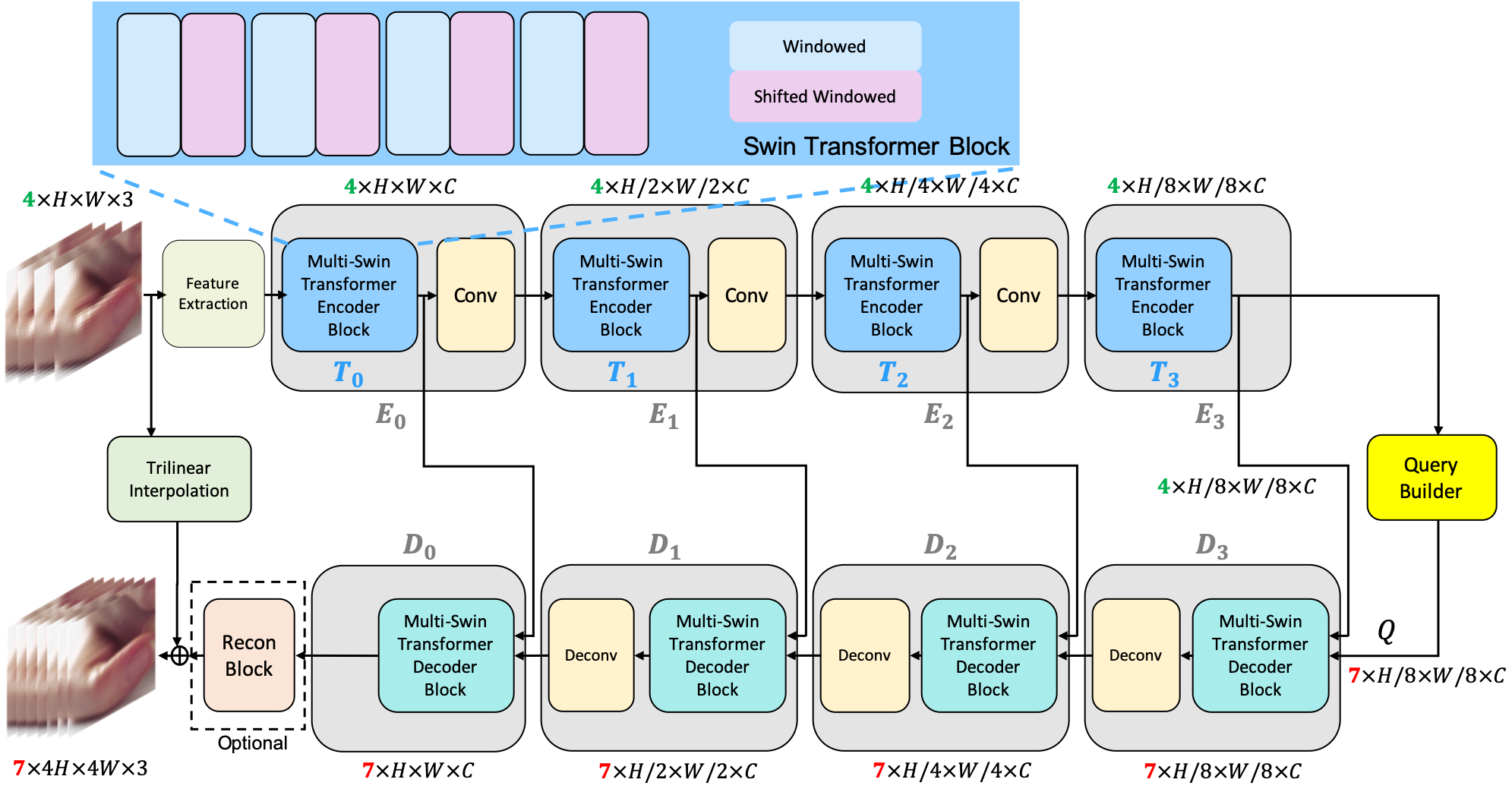} 
\end{center}
   \caption{\textbf{The architecture of the proposed RSTT.} The features extracted from four input LFR and LR frames are processed by encoders $E_k,k=0,1,2,3$ to build dictionaries that will be used as inputs for the decoders $D_k,k=0,1,2,3$. The query builder generates a vector of queries ${Q}$ which are then used to synthesize a sequence of seven consecutive HFR and HR frames. The Multi-Swin transformer encoder and decoder blocks contain a set of repeated Swin Transformer Blocks, which are illustrated in more details in Figure \ref{fig:swinEnc} and \ref{fig:swinDec}.}
\label{fig:arch}
\end{figure*}

\section{The proposed method}

In this section, we first give an overview of the proposed approach in~\Cref{subsec:overview}. Then we explain the encoder and decoder part of our spatial-temporal transformer in~\Cref{sec:enc} and~\Cref{sec:dec}, respectively. Finally, the training details are given in~\Cref{subsec:train}.

\subsection{Network overview}\label{subsec:overview}

Given $(n+1)$ LFR and LR  frames $\mathcal{I}^L=\{I_{2t-1}^L\}_{t=1}^{n+1}$ of size $H\times W\times 3$, a STVSR model generates $2n+1$ HFR and HR frames $\mathcal{I}^H=\{I_t^H\}_{t=1}^{2n+1}$ of size $ 4H\times 4W\times 3$, where $t$ denotes the time stamp of a frame. Note that only frames with odd time stamp in $\mathcal{I}^H$ has the LR counterparts in $\mathcal{I}^L $. 

We propose a cascaded UNet-style transformer, named \textbf{R}eal-time \textbf{S}patial \textbf{T}emporal \textbf{T}ransformer (\textbf{RSTT}), to interpolate the LR sequence $\mathcal{I}^L$ in both time and space simultaneously \emph{without having explicit separations of the model into spatial and temporal  interpolation modules}. One may shortly observe that this design is a distinct advantage over the existing CNN-based methods since it leads to a real-time inference speed while maintains similar performance.

We let $f$ denote the underlying function modeled by RSTT, which takes four consecutive LFR and LR frames in $\mathcal{I}^L$ and outputs seven HFR and HR frames in the sequence:
\begin{align}
\begin{split}
f &: (I_{2t-1}^L, I_{2t+1}^L, I_{2t+3}^L, I_{2t+5}^L) \\
&\mapsto
(I_{2t-1}^H, I_{2t}^H, I_{2t+1}^H, I_{2t+2}^H,I_{2t+3}^H,I_{2t+4}^H,I_{2t+5}^H)
\end{split}
\end{align}
As illustrated in~\Cref{fig:arch}, RSTT mainly consists of four encoders $E_k, k=0,1,2, 3$, and corresponding decoders $D_k, k=0,1,2,3$. In RSTT, a feature extraction block firstly extracts the features of the four input frames, denoted by
$
(F_{2t-1}^L, F_{2t+1}^L, F_{2t+3}^L, F_{2t+5}^L)
$; then, a multi-video Swin Transformer encoder block $\mathcal{T}_{\text{swin}}$ takes the features as input:
\begin{align*}
T_0 = \mathcal{T}_{\text{swin}}(F_{2t-1}^L, F_{2t+1}^L, F_{2t+3}^L, F_{2t+5}^L),
\end{align*}
where $T_0$ is the embedded feature generated by Swin Transformer.  
Let $\Phi$ denote the convolutional block in $E_0, E_1$ and $E_2$, one can write $E_0=\Phi(T_0)$. Subsequently, we have
\begin{align}\label{eq:tenc}
    \begin{cases}
    T_k = \mathcal{T}_{\text{swin}}(E_{k-1}), &\quad k=1,2,3 \\
    E_k = \Phi(T_k), &\quad k=1,2 \\
    E_3 = T_3
    \end{cases}
\end{align}
Note that each of the encoders $E_k,k=0,1,2,3$, has four output channels corresponding to the four time stamps of the input LFR and LR frames. To make it clear, we use
\begin{align*}
E_k\equiv (E_{k,2t-1}, E_{k,2t+1}, E_{k,2t+3},E_{k,2t+5} )
\end{align*}
to denote the four output feature maps of each $E_k$. In fact, a reusable dictionary is built in each $E_k$, and the details of the encoder architecture are presented in~\Cref{sec:enc}.

After computing $E_3$, RSTT constructs a query builder that generates features for interpolating HFR and HR frames at finer time stamps. Specifically, we define the query $Q$ as seven-channel feature maps with
\begin{align}\label{eq:q}
\begin{split}
Q:=\Big(E_{3,2t-1}, &\tfrac{1}{2}(E_{3,2t-1}+E_{3,2t+1}),E_{3,2t+1},
\\
&\tfrac{1}{2}(E_{3,2t+1}+E_{3,2t+3}),E_{3,2t+3},\\
&\tfrac{1}{2}(E_{3,2t+3}+E_{3,2t+5}), E_{3,2t+5}\Big)
\end{split}
\end{align}
As indicated in~\eqref{eq:q}, for odd HFR and HR frames which already have their LFR and LR counterparts, we just adopt the learnt features from the encoder $E_3$ as the queries; while for even frames that have \emph{no} LFR and LR counterparts, we use the mean features of their adjacent frames as the queries. 

We are now ready to synthesize the HFR and HR frames by feeding the decoders with the query and the outputs of encoders. As shown in~\Cref{fig:arch}, similar to~\eqref{eq:tenc}, we have
\begin{align}
    \begin{cases}
    D_3 = \Phi^{-1}(\mathcal{T}^{-1}_{\text{swin}}(T_3, Q)),  \\
    D_k = \Phi^{-1}(\mathcal{T}^{-1}_{\text{swin}}(T_k, D_{k+1})), &\quad k=1,2\\
    D_0 = \mathcal{T}^{-1}_{\text{swin}}(T_0, D_{1})
    \end{cases}
\end{align}
where $\mathcal{T}^{-1}_{\text{swin}}$ is the multi-video Swin Transformer decoder block and $\Phi^{-1}$ denotes the deconvolutional block in $D_1,D_2$ and $D_3$. The details of the decoder architecture are presented in~\Cref{sec:dec}. For the final synthesis, we learn the residuals instead of the HFR and HR frames themselves. We simply use a trilinear interpolation of the input frames to work as a warming start of the output frames. 

We remark that the key to the architecture of RSTT is the reusable dictionaries built in the encoders $E_k$ based on the input LFR and LR frames, which are then utilized in decoders ${D}_k$ combined with queries to synthesize the HFR and HR frames. This design is advantageous over the duplicate feature fusions appearing in many existing methods, \eg, deformable convolutions and ConvLSTM in Zooming SlowMo \cite{xiang2020zooming} and TMNet \cite{xu2021temporal}), and thus accelerates the inference speed to a large extent.

\begin{figure}[t]
\begin{center}
\includegraphics[width=0.47\textwidth]{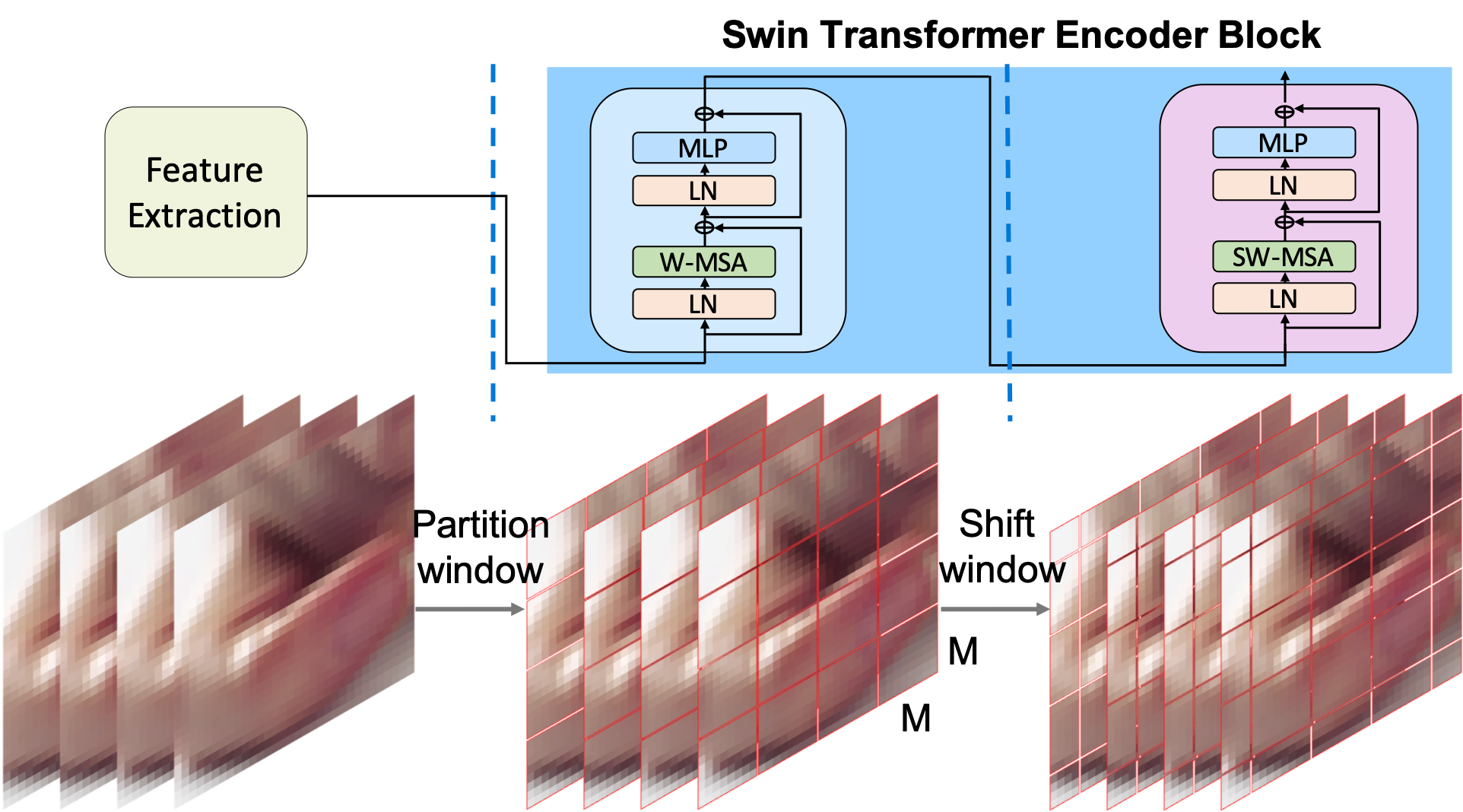} 
\end{center}
   \caption{\textbf{The basic Swin Transformer encoder block used in $\bm{ E_k,k=0,1,2,3}$ of RSTT; see Figure~\ref{fig:arch}.} It first computes multi-head self attentions in each window partition, and then in each shifted window partition. Here, {LN} stands for Layer Normalization, {W-MSA} is Windowed Multi-Head Self-Attention and {SW-MSA} is Shifted Windowed Multi-Head Self-Attention.}
\label{fig:swinEnc}
\end{figure}

\subsection{Encoder} \label{sec:enc}

In this subsection, we explain in details the encoder architecture of our RSTT. Before moving on, for the feature extraction module shown in~\Cref{fig:arch}, we use a single convolutional layer with kernel size $3\times3$ to extract $C$ features from four input LFR and LR RGB frames. This shallow feature extractor is significantly smaller than the five residual blocks used in Zooming SlowMo \cite{xiang2020zooming,xiang2021zooming} and TMNet~\cite{xu2021temporal}, and thus is computationally efficient.


Following the light-weight feature extractor, the encoder part of RSTT consists of four stages, denoted by $E_k,k=0,1,2,3$, each of which is a stack of Swin Transformer \cite{liu2021swin} blocks followed by a convolution layer (except $E_3$). Inside $E_k$, Swin Transformer blocks take the approach of shifting non-overlapping windows to reduce the computational cost while keeping the ability of learning long-range dependencies. As demonstrated in Figure \ref{fig:swinEnc}, given a pre-defined window size $M\times M$, a Swin Transformer block partitions the input video frames of size $N\times H\times W\times C$ 
into $N\times\lceil\frac{H}{M}\rceil\times\lceil\frac{W}{M}\rceil\times C$ 
non-overlapping windows, where we choose $N=4$, $M=4$ and  $C=96$ in our experiments. After flattening the features in each window to produce feature maps of size $\frac{NHW}{M^2}\times M^2 \times C$, Layer Normalization~({LN}) is applied to the features before Window-based Multi-head Self-Attention ({W-MSA}) \cite{liu2021swin} computes the local attention inside each window. Next, a Multi-Layer Perception~({MLP}) following another LN layer are used for further transformation. An additional Swin Transformer block with 
Shifted Window-based Multi-head Self-Attention ({SW-MSA}) \cite{liu2021swin} is then applied to 
introduce the cross-window connections.
In this second Swin Transformer block, every module is the same as the previous block except that the input features are shifted 
by $\lfloor\frac{M}{2}\rfloor\times\lfloor\frac{M}{2}\rfloor$ before window partitioning.
In this way, Swin Transformer blocks are able to reduce computational costs while capturing long-range dependencies along both the spatial and temporal dimension. Finally, the output of a stack of such Swin Transformer blocks are downsampled by a convolutional layer with stride of two, serving as the input of the next encoder stage and the decoder stage.

\begin{figure}[t]
\begin{center}
\includegraphics[width=0.47\textwidth]{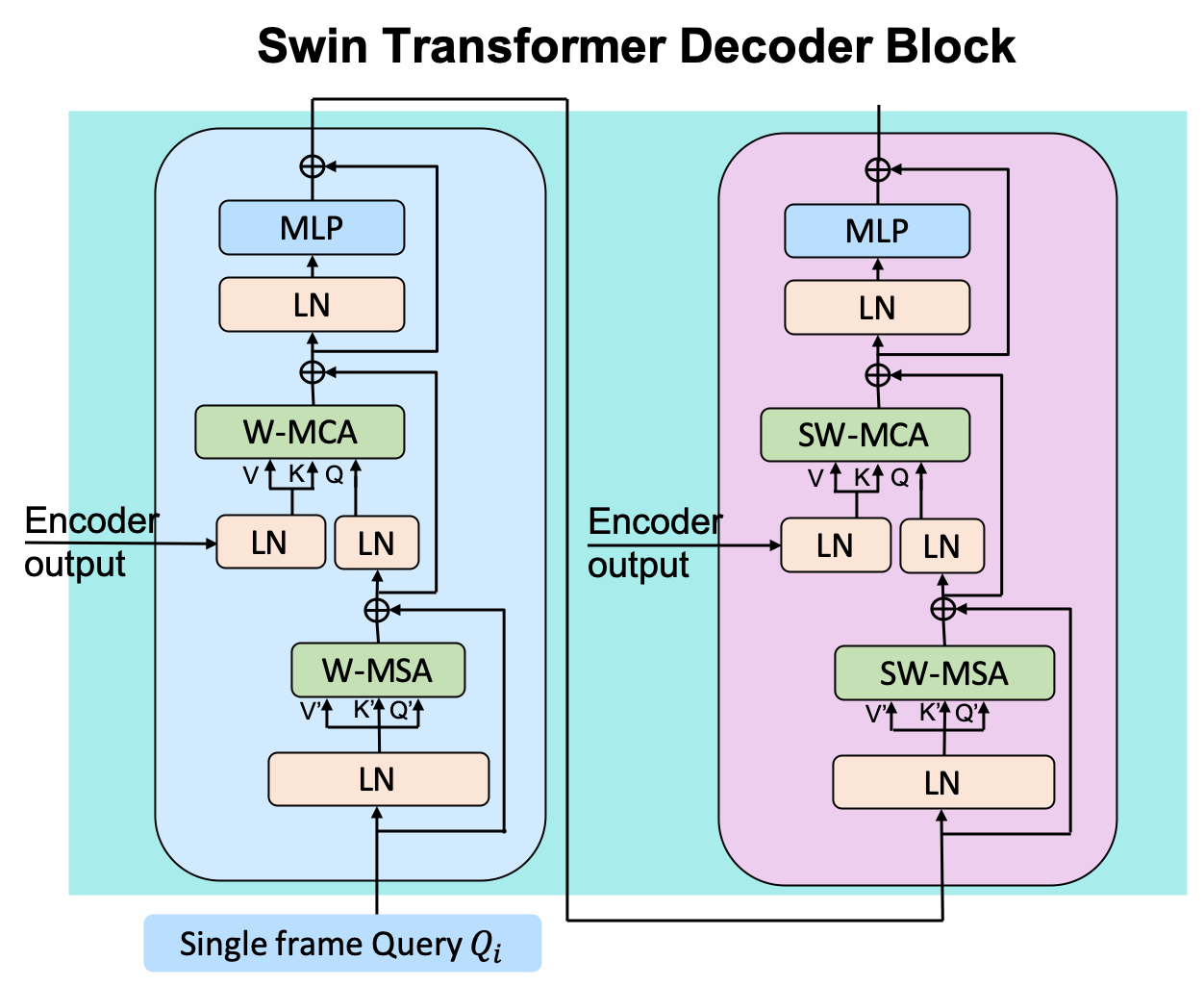} 
\end{center}
   \caption{\textbf{The basic Swin Transformer decoder block used in $\bm{D_k,k=0,1,2,3}$ of RSTT; see Figure \ref{fig:arch}.} It takes a query $Q$ and the output from the corresponding encoder $E_k$ as the input, and  outputs HFR and HR features for spatial-temporal interpolation. Here, {MCA} stands for Multi-Head Cross-Attention, and other notations are similar to those in~\Cref{fig:swinEnc}.}
\label{fig:swinDec}
\end{figure}

\begin{table*}
\footnotesize
\begin{center}
\begin{tabular}{@{}ccccccccccc@{}}
\toprule
Method & \multicolumn{2}{c}{Vid4}  & \multicolumn{2}{c}{Vimeo-Fast} & \multicolumn{2}{c}{Vimeo-Medium} & \multicolumn{2}{c}{Vimeo-Slow} &
FPS & Parameters \\
\cmidrule(lr){2-3} \cmidrule(lr){4-5} \cmidrule(lr){6-7} \cmidrule(lr){8-9}
VFI+(V)SR/STVSR & PSNR$\uparrow$ & SSIM$\uparrow$ & PSNR$\uparrow$ & SSIM$\uparrow$ & PSNR$\uparrow$ & SSIM$\uparrow$ & PSNR$\uparrow$ & SSIM$\uparrow$ & $\uparrow$ &  (Millions) $\downarrow$ \\
\hline\hline
SuperSloMo \cite{jiang2018super} + Bicubic & 22.84 & 0.5772 & 31.88 & 0.8793 & 29.94 & 0.8477 & 28.37 & 0.8102 & - & 19.8 \\
SuperSloMo \cite{jiang2018super} + RCAN \cite{zhang2018rcan} & 23.80  & 0.6397 & 34.52 & 0.9076 & 32.50 & 0.8884 & 30.69 & 0.8624 & - & 19.8+16.0 \\
SuperSloMo \cite{jiang2018super} + RBPN \cite{RBPN2019} & 23.76 & 0.6362  & 34.73 & 0.9108 & 32.79 & 0.8930 & 30.48 & 0.8584 & - &  19.8+12.7 \\
SuperSloMo \cite{jiang2018super} + EDVR \cite{wang2019edvr} & 24.40 & 0.6706 & 35.05 & 0.9136 & 33.85 & 0.8967 & 30.99 & 0.8673 & - &  19.8+20.7 \\
\hline
SepConv \cite{niklaus2017video_sepcov} + Bicubic & 23.51 & 0.6273 & 32.27 & 0.8890 & 30.61 & 0.8633 & 29.04 & 0.8290 & - &  21.7 \\
SepConv \cite{niklaus2017video_sepcov} + RCAN \cite{zhang2018rcan} & 24.92 & 0.7236 & 34.97 & 0.9195 & 33.59 & 0.9125 & 32.13 & 0.8967 & - &  21.7+16.0 \\ 
SepConv \cite{niklaus2017video_sepcov} + RBPN \cite{RBPN2019} & 26.08 & 0.7751 & 35.07 & 0.9238 & 34.09 & 0.9229 & 32.77 & 0.9090 & - &  21.7+12.7 \\
SepConv \cite{niklaus2017video_sepcov} + EDVR \cite{wang2019edvr} & 25.93 & 0.7792 & 35.23 & 0.9252 & 34.22 & 0.9240 & 32.96 & 0.9112 & - &  21.7+20.7 \\
\hline
DAIN \cite{bao2019depth} + Bicubic & 23.55 & 0.6268 & 32.41 & 0.8910 & 30.67 & 0.8636 & 29.06 & 0.8289 & - &  24.0 \\
DAIN \cite{bao2019depth} + RCAN \cite{zhang2018rcan} & 25.03 & 0.7261 & 35.27 & 0.9242 & 33.82 & 0.9146 & 32.26 & 0.8974 & - &  24.0+16.0 \\
DAIN \cite{bao2019depth} + RBPN \cite{RBPN2019} & 25.96 & 0.7784 & 35.55 & 0.9300 & 34.45 & 0.9262 & 32.92 & 0.9097 & - &  24.0+12.7 \\
DAIN \cite{bao2019depth} + EDVR \cite{wang2019edvr} & 26.12 & 0.7836 & 35.81 & 0.9323 & 34.66 & 0.9281 & 33.11 & 0.9119 & - &  24.0+20.7 \\
\hline
STARnet \cite{STAR2020} & 26.06 & 0.8046 & 36.19 & 0.9368 & 34.86 & 0.9356 & 33.10 & \textcolor{red}{\textbf{0.9164}} & 3.85 &  111.61 \\
Zooming SlowMo \cite{xiang2020zooming} & \textbf{26.31} & 0.7976 & \textcolor{blue}{\textbf{36.81}} & \textcolor{blue}{\textbf{0.9415}} & 35.41 & 0.9361 & 33.36 & 0.9138 & \textbf{15.59} & 11.10 \\
TMNet \cite{xu2021temporal} & \textcolor{red}{\textbf{26.43}} & \textcolor{red}{\textbf{0.8016}} & \textcolor{red}{\textbf{37.04}} & \textcolor{red}{\textbf{0.9435}} & \textbf{35.60} & \textcolor{blue}{\textbf{0.9380}} & \textcolor{red}{\textbf{33.51}} & \textcolor{blue}{\textbf{0.9159}} & 14.33 &  12.26 \\
\hline
RSTT-L & \textcolor{red}{\textbf{26.43}} & \textcolor{blue}{\textbf{0.7994}} & \textbf{36.80} & \textbf{0.9403} &  \textcolor{red}{\textbf{35.66}} & \textcolor{red}{\textbf{0.9381}} & \textcolor{blue}{\textbf{33.50}} & \textbf{0.9147} & 14.98 & \textbf{7.67} \\
RSTT-M & \textcolor{blue}{\textbf{26.37}} & \textbf{0.7978} & 36.78 & 0.9401 & \textcolor{blue}{\textbf{35.62}} & \textbf{0.9377} & \textbf{33.47} & 0.9143 & \textcolor{blue}{\textbf{19.07}} & \textcolor{blue}{\textbf{6.08}} \\
RSTT-S & 26.29 & 0.7941 & 36.58 & 0.9381 & 35.43 & 0.9358 & 33.30 & 0.9123 & \textcolor{red}{\textbf{26.19}} & \textcolor{red}{\textbf{4.49}} \\
\bottomrule
\end{tabular}
\end{center}
\caption{\textbf{Quantitative comparisons on various datasets with the state-of-the-art STVSR methods.} PSNR and SSIM are computed on Y channel only, as same as \cite{xiang2020zooming}. Top three numbers of each column are \textbf{bolded}, with the best in \textcolor{red}{\textbf{red}} and the second best in \textcolor{blue}{\textbf{blue}}. FPS is computed on Nvidia Quadro RTX 6000 machine and on Vid4 dataset, which has the output frame size of $720\times576$.} \label{QuantEva}
\end{table*}

\subsection{Decoder} \label{sec:dec}

Same as the encoder part, we use four stages of decoders followed by a deconvolutional layer for feature up-sampling. The decoders $D_k,k=0,1,2,3$ generate per-output-frame features in each level of details by repeatedly querying the dictionaries (the key-value pairs $(K, V)$ as shown in~\Cref{fig:swinDec}) constructed from the encoders $E_k$'s in the same level. Each decoder consists of several (the same number as its corresponding encoder) Swin Transformer blocks, and each of the blocks takes {two} inputs: one is the output features from the encoder and the other is a single frame query, as shown in~\Cref{fig:arch} and Figure \ref{fig:swinDec}. 

In RSTT, the first stage query $Q = \{Q_i\}_{i=1}^7$ for $D_3$ is interpolated from the the last encoder $E_3$ (see~\eqref{eq:q}) and the later queries are the outputs of the previous decoders. To generate seven HFR and HR output frames, each Swin Transformer decoder block queries the dictionary seven times. As a result, for a decoder contains $S$ such blocks, we need to query $7S$ times.  In practice, the query is performed by Windowed Multi-Head Cross-Attention (W-MCA) \cite{liu2021swin} and its shifted version (SW-MCA) \cite{liu2021swin} (see Figure \ref{fig:swinDec}). Note that only the first Swin Transformer decoder block uses $Q$ as query while the rest $S-1$ blocks use the output of the previous block as query. Importantly, dictionaries provided by the encoders are pre-computed for reuse in each block. Suppose we have three Swin Transformer decoder blocks in each $D_k$, the spatial-temporal dictionaries built from the encoders are queried (reused) for $7\times3=21$ times, which is advantageous over the duplication of future fusions in the existing approaches. This design is both computationally efficient and helpful in reducing the model size.


\myparagraph{Final reconstruction module.}
The output features of the last decoder $D_0$ can be further processed by an optional reconstruction module to generate the final frames (see Figure~\ref{fig:arch}). We use a module consisting of a 1-to-4 PixelShuffle operation and a single convolutional layer. This design is much more light-weight compared to the practices adopted in Zooming SlowMo~\cite{xiang2020zooming} and TMNet~\cite{xu2021temporal}, both of which use $40$ residual blocks to perform the spatial super resolution. 
We compare the performance of RSTT with and without such spatial reconstruction module in Section \ref{sec:AS}.

\begin{figure*}[!ht]
{\footnotesize \hspace{0.7in}Ground-truth\hspace{1.1in} Ground-truth \hspace{0.6in}StarNet~\cite{STAR2020}\hspace{0.3in}Zooming SlowMo~\cite{xiang2020zooming}\hspace{0.4in}TMNet~\cite{xu2021temporal}} \\
    \centering
    \includegraphics[width=0.99\textwidth]{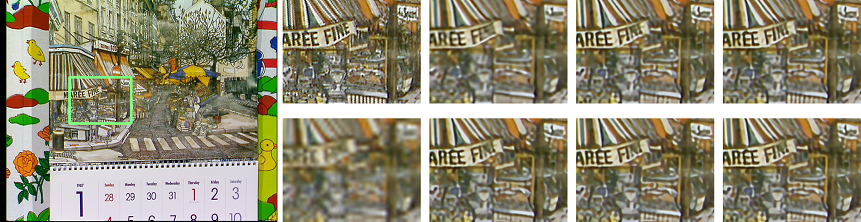}\\
\vspace{-1in}
{\footnotesize \hspace{2.1in} Overlaid \hspace{0.7in}RSTT-S\hspace{0.8in}RSTT-M\hspace{0.8in}RSTT-L} \\
\vspace{0.9in}
{\footnotesize \hspace{0.6in}Ground-truth\hspace{1.1in} Ground-truth \hspace{0.6in}StarNet~\cite{STAR2020}\hspace{0.3in}Zooming SlowMo~\cite{xiang2020zooming}\hspace{0.4in}TMNet~\cite{xu2021temporal}} \\
    \includegraphics[width=0.99\textwidth]{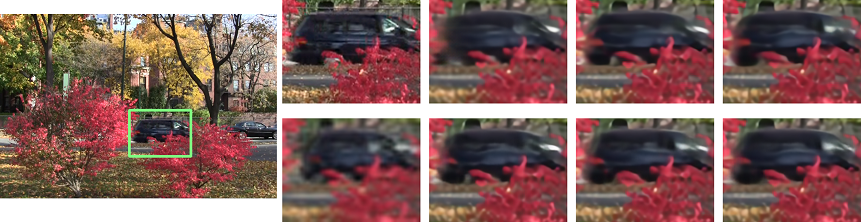}\\
\vspace{-1in}
{\footnotesize \hspace{2.1in} Overlaid \hspace{0.7in}RSTT-S\hspace{0.8in}RSTT-M\hspace{0.8in}RSTT-L} \\
\vspace{0.9in}
{\footnotesize \hspace{0.6in}Ground-truth\hspace{1.1in} Ground-truth \hspace{0.6in}StarNet~\cite{STAR2020}\hspace{0.3in}Zooming SlowMo~\cite{xiang2020zooming}\hspace{0.4in}TMNet~\cite{xu2021temporal}} \\
    \includegraphics[width=0.99\textwidth]{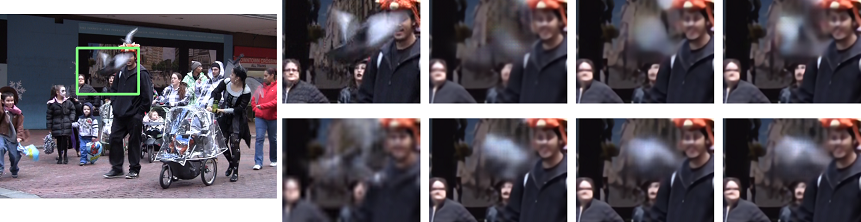}\\
\vspace{-1in}
{\footnotesize \hspace{2.1in} Overlaid \hspace{0.7in}RSTT-S\hspace{0.8in}RSTT-M\hspace{0.8in}RSTT-L} \\
\vspace{1in}
    \caption{\textbf{Visual comparisons on the Vid4 dataset~\cite{liu2011bayesian}.} RSTT with three different sizes of architectures achieve the state-of-the-art performance in terms of visual qualities on various scenarios.}
    \label{fig:visual}
\end{figure*}

\subsection{Training details}\label{subsec:train}

We train the proposed RSTT model using Adam with $\mathcal{L}_2$ and decoupled weight decay \cite{loshchilov2018decoupled} by setting $\beta_1=0.9$ and $\beta_2=0.99$. The initial learning rate is set to $2\times 10^{-4}$ and is gradually decayed following the scheme of Cosine annealing with restart \cite{gotmare2018closer} set to $10^{-7}$. The restart performs at every 30,000 iterations. We train our model on two Nvidia Quadro RTX 6000 with batch size set to 7$\sim$10, depending on the particular model architecture. 

\myparagraph{Objective function.} The Charbonnier loss is computed between the estimated frame $I^H $ and the ground truth $\hat{I}^H$:
\begin{equation*}
   L(\hat{I}^H, I^H) = \sqrt{\lVert\hat{I}^H - I^H\rVert^2 + \epsilon^2},
\end{equation*}
where $\epsilon$ is set to $10^{-3}$ in our experiments.

\myparagraph{Training dataset.} We train our models on Vimeo-90K~\cite{xue2019video}, which contains over 60,000 seven-frame video sequences. Many state-of-the-art methods~\cite{xiang2020zooming,xiang2021zooming,xu2021temporal} also use this dataset for training. For Vimeo-90K, the input LFR and LR frames are four frames of size $112\times64$, and the output HFR and HR frames are seven frames of size $448\times256$ (exactly $4\times$ larger in both height and width). 

\myparagraph{Evaluation.} The models are evaluated on Vid4 \cite{liu2011bayesian} and Vimeo-90K~\cite{xue2019video} datasets. Vid4 is a small dataset consists of four video sequences of different scenes with $180\times144$ input frames and $720\times576$ output frames. Vimeo-90K validation set is split into fast, medium and slow motion sets as in~\cite{xiang2020zooming} that contains 1225, 4972 and 1610 video clips. 

\section{Experiments}

We train the proposed RSTT for three versions with small (S), medium (M) and large (L) architectures, corresponding to the number of Swin Transformer blocks used in each stage of encoder $E_k$ and decoder $D_k$ set to 2, 3 and 4, respectively. We term the three models as RSTT-S, RSTT-M and RSTT-L, and then compare them with other existing methods quantitatively and qualitatively.


\subsection{Quantitative evaluation} 

We use Peak Signal to Noise Ratio (PSNR) and Structural Similarity (SSIM) as evaluation metrics for quantitative comparison. We also compare the model inference time in Frame Per Second (FPS) and model size in terms of the number of parameters, as shown in Table \ref{QuantEva}. We do not list the FPS of methods that sequentially apply separated VFI and VSR models, since they are much slower than the other competitors, as reported in \cite{xiang2020zooming,xu2021temporal}. 

\begin{table*}[ht]
\footnotesize
\begin{center}
\begin{tabular}{@{}ccccccccccc@{}}
\toprule
Method & \multicolumn{2}{c}{Vid4}  & \multicolumn{2}{c}{Vimeo-Fast} & \multicolumn{2}{c}{Vimeo-Medium} & \multicolumn{2}{c}{Vimeo-Slow} &
FPS & Parameters \\
\cmidrule(lr){2-3} \cmidrule(lr){4-5} \cmidrule(lr){6-7} \cmidrule(lr){8-9}
VFI+(V)SR/STVSR & PSNR$\uparrow$ & SSIM$\uparrow$ & PSNR$\uparrow$ & SSIM$\uparrow$ & PSNR$\uparrow$ & SSIM$\uparrow$ & PSNR$\uparrow$ & SSIM$\uparrow$ & $\uparrow$ &  (Millions) $\downarrow$ \\
\hline\hline
RSTT-M-Recon & \textcolor{red}{\textbf{26.37}} & \textcolor{red}{\textbf{0.7988}} & \textcolor{red}{\textbf{36.80}} & \textcolor{blue}{\textbf{0.9400}} & \textcolor{red}{\textbf{35.66}} & \textcolor{red}{\textbf{0.9381}} & \textcolor{red}{\textbf{33.58}} & \textcolor{red}{\textbf{0.9160}} & 17.02 & 7.74 \\
RSTT-M & \textcolor{red}{\textbf{26.37}} & \textcolor{blue}{\textbf{0.7978}} & \textcolor{blue}{\textbf{36.78}} & \textcolor{red}{\textbf{0.9401}} & \textcolor{blue}{\textbf{35.62}} & \textcolor{blue}{\textbf{0.9377}} & \textcolor{blue}{\textbf{33.47}} & \textcolor{blue}{\textbf{0.9143}} & \textbf{19.07} & \textcolor{blue}{\textbf{6.08}} \\
RSTT-S-Recon & \textcolor{blue}{\textbf{26.29}} & \textbf{0.7951} & 36.56 & 0.9376 & \textbf{35.45} & \textbf{0.9361} & \textbf{33.40} & \textbf{0.9140} & \textcolor{blue}{\textbf{22.62}} & \textbf{6.15} \\
RSTT-S & \textcolor{blue}{\textbf{26.29}} & 0.7941 & \textbf{36.58} & \textbf{0.9381} & 35.43 & 0.9358 & 33.30 & 0.9123 & \textcolor{red}{\textbf{26.19}} & \textcolor{red}{\textbf{4.49}} \\
\bottomrule
\end{tabular} 
\end{center}
\caption{\textbf{Quantitative comparisons of RSTT with and without the spatial reconstruction block.} Top-three numbers of each column are \textbf{bolded}, with the best in \textcolor{red}{\textbf{red}} and the second best in \textcolor{blue}{\textbf{blue}}.}
\label{tab:ablation}
\vspace{-.05in}
\end{table*} 

We observe that all the RSTT models achieve state-of-the-art performance in both Vid4 and Vimeo-90K datasets with significantly smaller model size and substantially higher inference speed. Moreover, the performance grows steadily with increasing number of Swin Transformer blocks stacked in the architecture, from RSTT-S, -M to -L. Specifically, in Table \ref{QuantEva}, one can see that the smallest model RSTT-S performs similarly as Zooming SlowMo \cite{xiang2020zooming}, while RSTT-M outperforms Zooming SlowMo \cite{xiang2020zooming} in Vid4, Vimeo-Medium and Vimeo-Slow with significantly smaller number of parameters and faster inference speed. Our largest model RSTT-L outperforms TMNet \cite{xu2021temporal} on Vimeo-Medium, which is the largest dataset in Table \ref{QuantEva}, with $40\%$ smaller model size. We remark that our RSTT-S achieves a real-time rendering speed (more than 26 FPS) without sacrificing much performance.

\subsection{Qualitative evaluation}

We visually compare RSTT with other state-of-the art STVSR methods in Figure \ref{fig:visual}. We choose three different scenarios for the purpose of illustration:
\begin{itemize}[leftmargin=*]
    \item The first row shows the video of a \emph{still} calendar in front of a \emph{moving} camera. We observe that RSTT-S recovers details around the character. Texture details look more apparent compared with the result of StarNet \cite{STAR2020}.
    
    \item The second row shows the video taken by a \emph{still} camera in the wild with fast \emph{moving} vehicles. It is clear that RSTT outperforms StarNet~\cite{STAR2020} and Zooming SlowMo~\cite{xiang2020zooming} with better contours of the moving vehicle.
    
    \item The third row illustrates a difficult case, where \emph{both} the camera and the foreground objects are \emph{moving}, especially the fast-flying pigeon. From the overlaid view, one can see that the pigeons in consecutive frames are barely overlapped. Our models give relatively better motion interpolations in this case compared with other state-of-the-arts. In addition, with the increasing sizes of our models, from RSTT-S to RSTT-L, we observe better interpolations. 
\end{itemize}

\subsection{Advantages of RSTT} \label{sec:AS}

We analyze the effectiveness of the Swin Transformer blocks used in encoders and decoders by comparing with using an optional spatial reconstruction block in Figure~\ref{fig:arch} with $10$ residual blocks (see Table~\ref{tab:ablation}). This block is similar to but smaller than the $40$ residual blocks used in Zooming SlowMo \cite{xiang2020zooming} and TMNet \cite{xu2021temporal}. We observe that the additional reconstruction block only slightly changes the evaluation results. There are hardly any differences in performance ($\pm${0.02}db in psnr) on Vid4, Vimeo-Fast and Vimeo-Medium datasets between RSTT models with and without adding reconstruction blocks. However, both the inference time and the network size are largely increased. Furthermore, {RSTT-M} ({6.08}M parameters) exhibits non-negligible improvement over {RSTT-S-Recon} ({6.15}M parameters) in all of the datasets ({$\geq$0.2}db in PSNR on Vimeo-Fast) with an even smaller model size. This reveals the effectiveness of our design, indicating larger spatial reconstruction block is unnecessary to RSTT. Note that we do not train a model with such additional block on RSTT-L due to the limited time, but we believe a similar pattern holds.

\subsection{Limitations of RSTT}

\myparagraph{Long training time.} Like other transformer-based methods~\cite{dosovitskiy2021an}, the required training time of RSTT is relatively long. It takes more than twenty-five days for convergence with the usage of two Nvidia Quadro RTX 6000 cards.  

\myparagraph{Lack the flexibility to interpolate at arbitrary time stamps.} Unlike TMNet \cite{xu2021temporal}, RSTT lacks the flexibility of interpolating an intermediate frame at arbitrary time stamps since the Query $Q$ defined by \eqref{eq:q} is fixed. However, we remark that this can be achieved by slightly rephrasing Query $Q$ for Decoder $D_3$. 
Suppose we would like to interpolate $n-1$ frames (at $n-1$ time stamps) between two frames, e.g., $E_{3,2t-1}$ and $E_{3,2t+1}$, we just need to make queries on $\{\frac{i}{n}E_{3,2t-1}+(1-\frac{i}{n})E_{3,2t+1}\}_{i=1}^{n-1}$ instead of $\frac{1}{2}E_{3,2t-1}+\frac{1}{2}E_{3,2t+1}$ where $n=2$ as a special case in \eqref{eq:q}. One might need to retrain the model to adopt such modifications, and we leave it as future work.

\section{Conclusion}

We presented a real-time spatial-temporal transformer (RSTT) for generating HFR and HR videos from LFR and LR ones. We considered to solve the space-time video super-resolution problem with a unified transformer architecture without having explicit separations of temporal and spatial sub-modules. Specifically, LFR and LR spatial-temporal features extracted from different levels of encoders are used to build dictionaries, which are then queried many times in the decoding stage for interpolating HFR and HR frames simultaneously. {We emphasize that the key innovation of the work is the novel {holistic formulation} of self-attentions in encoders and cross-attentions in decoders.} This holistic design leads to a significantly smaller model with much faster (real-time) inference speed compared with the state-of-the-art methods without noticeable difference in model performance. 

Future directions along this line include but are not limited to: fusions of dictionaries built in different levels of encoders to make computations more efficient; controllable temporal super-resolution with the flexibility to interpolate frames at arbitrary time stamps; and sophisticated training loss functions that helps to improve the visual quality. 



\newpage

\section{Supplementary Materials}

\subsection{Attention visualization}

\noindent
\begin{figure}[!htbp]
\begin{center}
 \includegraphics[width=.8\linewidth]{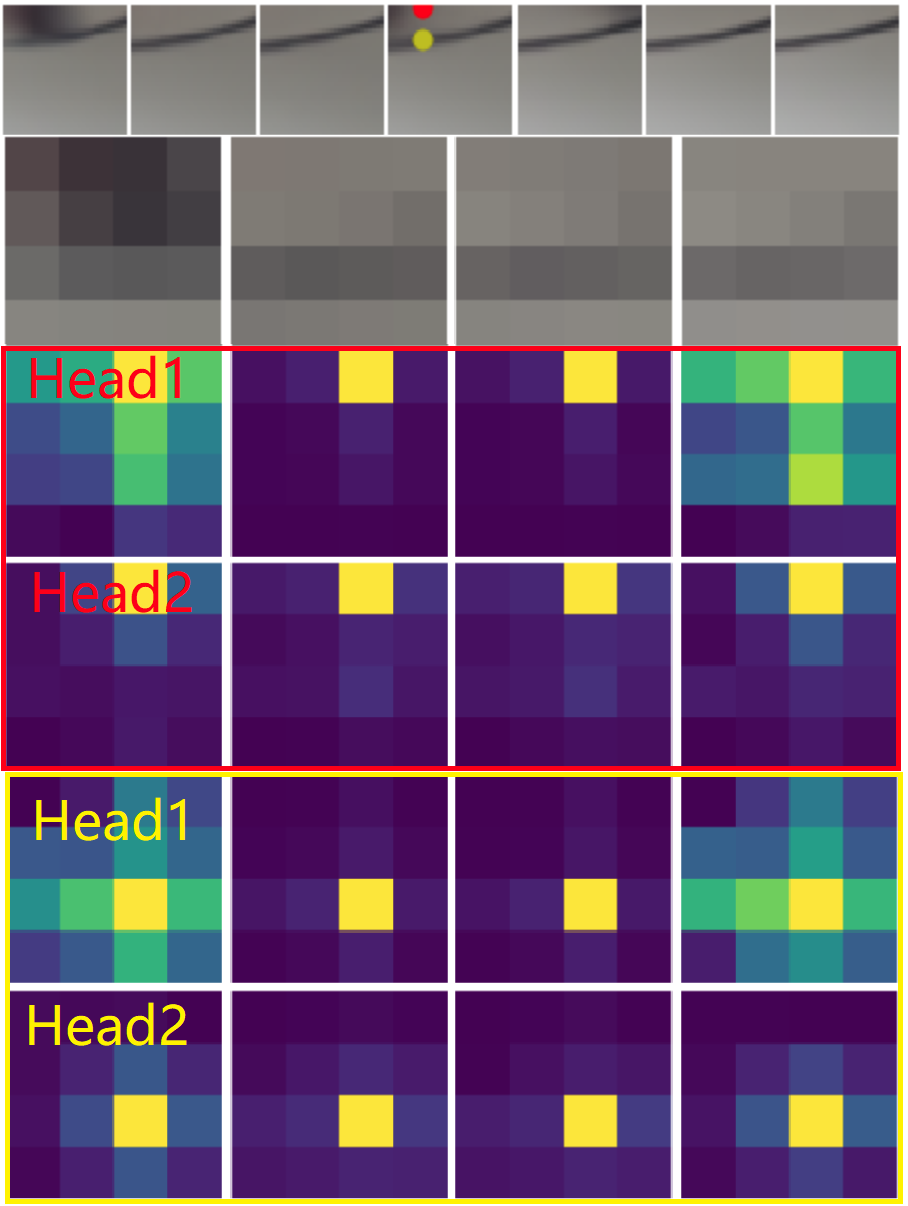}
\end{center}
\caption{Windowed 2-Head attention of the last decoder.} \label{attention}
\end{figure}

To better understand the mechanism of RSTT, We show windowed 2-Head cross attentions of the last decoder on a sequence in Vimeo-Fast at two different locations (see Figure \ref{attention}). Images bounded within red (yellow) box correspond to the red (yellow) dot highlighted in the outputs. We observe that the learnt attentions generally capture the local image structures in the finest level of detail.

\subsection{Effectiveness of Multi-head Cross Attention}

\begin{table}[!htbp]
\footnotesize
\begin{center}
\begin{tabular}{@{}cccccc@{}}
\toprule
Decoder & \multicolumn{2}{c}{Vid4}  & \multicolumn{2}{c}{Vimeo-Fast} \\
\cmidrule(lr){2-3} \cmidrule(lr){4-5} 
 & PSNR$\uparrow$ & SSIM$\uparrow$ & PSNR$\uparrow$ & SSIM$\uparrow$ \\
\hline\hline
MCA & 26.29 & 0.7941 & 36.58 & 0.9381 \\
Concat & 26.18 & 0.7879 & 36.29 & 0.9346 \\
Add & 26.13 & 0.7865 & 36.25 & 0.9340 \\
\bottomrule
\end{tabular}
\end{center}
\caption{\textbf{Quantitative comparisons on Vid4 and Vimeo-Fast datasets between MCA and other information fusion methods.} PSNR and SSIM are computed on Y channel only.} \label{MCA1}
\end{table}

\begin{table}[!htbp]
\footnotesize
\begin{center}
\begin{tabular}{@{}cccccc@{}}
\toprule
Decoder & \multicolumn{2}{c}{Vimeo-Medium} & \multicolumn{2}{c}{Vimeo-Slow} \\ \cmidrule(lr){2-3} \cmidrule(lr){4-5}
 & PSNR$\uparrow$ & SSIM$\uparrow$ & PSNR$\uparrow$ & SSIM$\uparrow$ \\
\hline\hline
MCA & 35.43 & 0.9358 & 33.30 & 0.9123 \\
Concat & 35.20 & 0.9332 & 33.15 & 0.9099 \\
Add & 35.15 & 0.9327 & 33.07 & 0.9090 \\
\bottomrule
\end{tabular}
\end{center}
\caption{\textbf{Quantitative comparisons on Vimeo-Medium and Vimeo-Slow datasets between MCA and other information fusion methods.} PSNR and SSIM are computed on Y channel only.} \label{MCA2}
\end{table}

RSTT employs Multi-head Cross Attention (MCA) to generate features in Decoders based on corresponding Encoders and Queries. To further investigate the effectiveness of this design, we replace MCA in RSTT-S with other ways of information combinations: concatenation and addition. The performance comparisons are shown in Table \ref{MCA1} and \ref{MCA2}. 

Here, both Windowed-MCA and Shifted Windowed-MCA are replaced by feature concatenation or feature addition by simply using 1d convolutions to match the feature sizes of stacked queries and the encoder outputs before this information fusion. Metrics in both Table \ref{MCA1} and \ref{MCA2} clearly testify the effectiveness of MCA over feature concatenation and feature addition.


{\small
\bibliographystyle{ieee_fullname}
\bibliography{egbib}
}

\end{document}